\documentclass[12pt]{article}
\usepackage{fullpage}
\usepackage[authoryear]{natbib}

\usepackage{type1cm}        
\usepackage{makeidx}         
\usepackage{graphicx}        
\usepackage{multicol}        
\usepackage[bottom]{footmisc}

\usepackage{newtxtext}       %
\usepackage[varvw]{newtxmath}       
\usepackage{algorithm,algorithmic}

\usepackage{amsthm}
\theoremstyle{definition}
\newtheorem{definition}{Definition}[section]

\newtheorem{theorem}{Theorem}

\usepackage{hyperref}
\usepackage{amsmath}
\usepackage{mathtools}
\usepackage{bm}
\usepackage{multicol}
\usepackage{multirow}
\usepackage[figuresright]{rotating}
\usepackage{siunitx}
\usepackage{booktabs}
\usepackage[]{graphicx}
\usepackage{subfigure}
\usepackage{graphicx}
\usepackage{float}
\usepackage{rotfloat}
\usepackage{caption}

\usepackage{algorithm}
\usepackage{algorithmic}

\makeindex             

\makeatletter
\newcommand{\specificthanks}[1]{\@fnsymbol{#1}}
\makeatother

\begin{document}

\title{Reinforcement Learning from Bagged Reward}


\author{
\normalsize
Yuting Tang\textsuperscript{1,2*}, 
Xin-Qiang Cai\textsuperscript{1*}, 
Yao-Xiang Ding\textsuperscript{3}, 
Qiyu Wu\textsuperscript{1},
Guoqing Liu\textsuperscript{4},\\
\normalsize
Masashi Sugiyama\textsuperscript{2,1} \\\\
\small
\textsuperscript{1}The University of Tokyo, Japan.\\
\small
\textsuperscript{2}RIKEN Center for Advanced Intelligence Project, Japan.\\
\small
\textsuperscript{3}State Key Lab for CAD \& CG, Zhejiang University, China.\\
\small
\textsuperscript{4}Microsoft Research AI4Science, China.
}

\date{}

\maketitle
\renewcommand{\thefootnote}{\fnsymbol{footnote}} 
\footnotetext[1]{Equal contribution.}

\begin{abstract}
In Reinforcement Learning (RL), it is commonly assumed that an immediate reward signal is generated for each action taken by the agent, helping the agent maximize cumulative rewards to obtain the optimal policy. However, in many real-world scenarios, designing immediate reward signals is difficult; instead, agents receive a single reward that is contingent upon a partial sequence or a complete trajectory. In this work, we define this challenging problem as \textit{RL from Bagged Reward} (RLBR), where sequences of data are treated as \textit{bags} with non-Markovian bagged rewards, leading to the formulation of Bagged Reward Markov Decision Processes (BRMDPs). Theoretically, we demonstrate that RLBR can be addressed by solving a standard MDP with properly redistributed bagged rewards allocated to each instance within a bag. Empirically, we find that reward redistribution becomes more challenging as the bag length increases, due to reduced informational granularity. Existing reward redistribution methods are insufficient to address these challenges. Therefore, we propose a novel reward redistribution method equipped with a bidirectional attention mechanism, enabling the accurate interpretation of contextual nuances and temporal dependencies within each bag. We experimentally demonstrate that our proposed method consistently outperforms existing approaches. The code is available at an anonymous link: \url{https://anonymous.4open.science/r/RLBR-F66E/}.
\end{abstract}

\section{Introduction}
\label{sec_introduction}
Reinforcement Learning (RL) has achieved remarkable success in various domains, including autonomous driving~\cite{kiran2021deep}, healthcare~\cite{yu2021reinforcement}, complex game playing~\cite{silver2016mastering, wurman2022outracing}, and financial trading~\cite{yang2020deep}. One common and essential assumption for most RL algorithms is the availability of immediate reward feedback at each time step of the decision-making process. However, designing such immediate reward feedback is quite difficult and may mislead the policy learning in many real-world applications~\cite{kwonreward,lee2023rlaif}. Recognizing this gap, numerous studies~\cite{watkins1989learning, ke2018sparse, gangwani2020ircr, ren2021rrd, zhang2023interpretable} have explored the concept of RL with Trajectory Feedback (RLTF), primarily focusing on trajectory feedback where rewards are allocated at the end of a sequence. 

\begin{figure}
  \centering
  \includegraphics[width=0.8\textwidth]{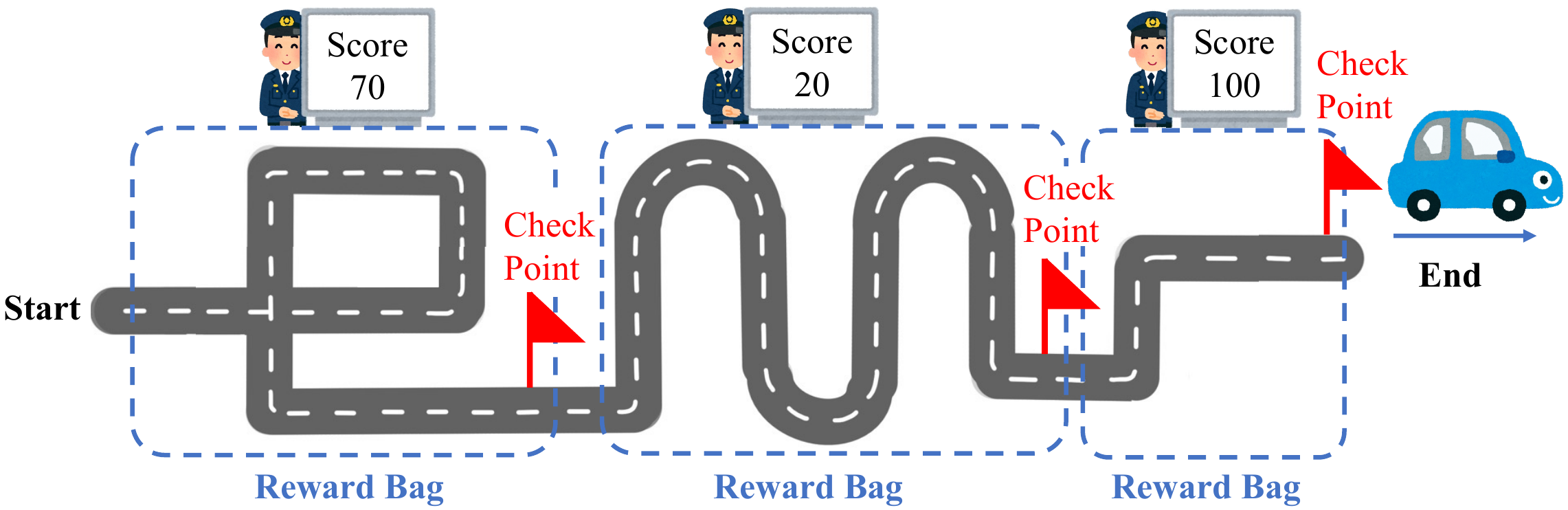}
  \caption{An example of the reward bag structure on an autonomous driving trajectory. Each segment of the driving sequence is evaluated and assigned a score by an evaluator.}
  \label{fig:example}
\end{figure}

However, real-world applications, such as autonomous driving (see Fig.~\ref{fig:example}), often feature complex non-immediate reward structures that RLTF cannot fully capture.
Providing rewards for every action is impractical, while focusing only on end goals ignores crucial aspects of the journey. Typically, rewards are linked to completing specific tasks or sequences of actions, rather than individual actions or the final objective \cite{early2022non, gaon2020reinforcement}. 
Previous studies focused on learning desirable policies with immediate rewards~\citep{712192, vinyals2019grandmaster} or trajectory feedback~\citep{arjona2019rudder} are not ideal methods for such scenarios.

To address these challenges, we introduce \textit{RL from Bagged Rewards} (RLBR), a framework that better aligns with real-world scenarios by considering the cumulative effect of a series of instances. In RLBR, sequences of instances are defined as \textit{bags}, each associated with a \textit{bagged reward}. This framework encompasses both the traditional RL setting, where each bag contains only a single instance, and the trajectory feedback setting, where a bag spans the entire trajectory, as special cases.
Furthermore, RLBR offers the potential to reduce the labeling workload by lessening the frequency of reward annotations. However, this benefit is balanced by increased learning complexity due to the reduced granularity of information.

In the RLBR framework, our focus is on leveraging bagged reward information to discern the significance of each instance within a bag and to understand the relationships among different bags. 
The challenge lies in accurately interpreting the contextual nuances within individual bags, as the instances within a bag are time-dependent on each other and their contributions to the bagged reward vary. 
Given the importance of context in RLBR, we turn to the bidirectional attention mechanism~\citep{seo2016bidirectional, vaswani2017attention, DBLP:conf/naacl/DevlinCLT19}, renowned for its effectiveness in contextual understanding, especially for time-dependent data. 
Specifically, we propose a Transformer-based reward model, leveraging the bidirectional attention mechanism to adeptly interpret context within a bag and allocate rewards to each instance accurately. This model can be utilized to enhance general RL algorithms, such as Soft Actor-Critic (SAC) \cite{haarnoja2018soft}, for environments with bagged rewards.

Our research contributes to the field in several ways. First, we establish the RLBR framework as a general problem setting by formulating it with an extension of the traditional Markov Decision Processes (MDPs), the \textit{Bagged Reward MDPs} (BRMDPs), as elaborated in Section 3.
In Section 4, we theoretically connect MDPs after reward redistribution to the original BRMDPs, which motivates us to employ a bidirectional attention mechanism within a Transformer-based framework to assign rewards to individual instances while capturing environmental dynamics.
Additionally, we propose an algorithm that alternates between optimizing this reward redistributing model and the policy, thereby enhancing the effectiveness of both components. 
Finally, in Section 5, our experiments show that the performance of baseline methods drops as the bag length increases, indicating that the larger bag length will increase the learning difficulty. Furthermore, we experimentally demonstrate the superiority of the proposed method through comparative performance analyses and validate the ability of the proposed model to mimic the reward distribution of the ground truth MDP, highlighting its contextual understanding and adaptability to environmental dynamics.

\section{Related Work}

With the growing attention on RLTF, where trajectory feedback is often considered the sum of rewards over the entire trajectory, new methods specifically designed for RLTF have emerged. 
Return Decomposition for Delayed Rewards (RUDDER) \cite{arjona2019rudder} used a return-equivalent formulation for precise credit assignment, with advancements incorporating expert demonstrations and language models \cite{liu2019sequence, widrich2021modern, patil2022align}. 
Both Iterative Relative Credit Refinement (IRCR) \cite{gangwani2020ircr} and Randomized Return Decomposition (RRD) \cite{ren2021rrd} assumed that each state-action pair contributes equally to the trajectory feedback. Specifically, IRCR presented a uniform reward redistribution model, while RRD proposed a novel upper bound for return-equivalent assumptions, integrating return decomposition with uniform reward redistribution. 
Additionally, \citet{han2022hc} modified RL algorithms to use sequence-level information, helping agents learn from broader structures and long-term outcomes. On the other hand, some studies on sparse rewards, such as reward shaping \cite{ng1999policy, tambwekar2019controllable, hu2020learning} and intrinsic reward methods \cite{sorg2010reward, pathak2017curiosity, zheng2018learning, zheng2021episodic}, could be directly applied to RLTF. However, these approaches were primarily focused on addressing the exploration-exploitation trade-off in sparse reward settings, which made them less ideal for solving RLTF problems.
A detailed discussion on the sparse reward setting and methods is provided in ``Discussion on Sparse Rewards'' in Appendix.

However, previous RLTF works that assume each state-action pair contributes equally to the trajectory feedback did not consider that instances contributing more to the feedback should receive greater attention. On the other hand, methods that consider sequence information struggled to capture effective information from long trajectories, resulting in suboptimal training outcomes, as evidenced by prior studies~\cite{ren2021rrd, zhang2023interpretable}.
Building on previous methods, we adopt a reward redistribution learning strategy to enhance policy learning in the context of reward bags. The proposed method aims to extract information from bag-level rewards, taking into account the contribution of each instance, ultimately learning an effective policy. Experimental results (see Section 5) demonstrate that the proposed method can effectively capture information from long sequences, and the rich information from the bags makes it more effective.

\section{Reinforcement Learning from Bagged Reward}
\label{sec_problem}

In this section, we first provide preliminaries on RL with immediate rewards and trajectory feedback, and then formulate the RLBR problem with an extension of the traditional MDPs, the BRMDPs.

\subsection{Preliminaries}
In traditional RL problems, which are modeled using finite MDPs, each state-action pair is promptly associated with a reward~\citep{712192}. This paradigm is encapsulated in a tuple \( \mathcal{M} = (\mathcal{S}, \mathcal{A}, \mathit{P}, \mathit{r}, \mu) \), with \( s\in \mathcal{S} \) and \(a\in \mathcal{A} \) as the sets of states and actions, \( \mathit{P} \) as the state transition probability function, \( \mathit{r} \) as the immediate reward function, and \( \mu \) as the initial state distribution. The primary objective in this framework is to discover a policy $\pi = p(a|s)$ that maximizes the cumulative sum of rewards over a horizon length $T$:
\begin{equation}
    J(\pi) = \mathbb{E}_{\pi, P, \mu} \bigg[ \sum_{t=0}^{T-1} r(s_t, a_{t}) \bigg].
\end{equation}

Given that associating each state-action pair with a reward is challenging in real-world scenarios and involves significant labeling costs, RLTF, also known as episodic or delayed rewards in some works, has become increasingly prominent in many applications~\citep{watkins1989learning,arjona2019rudder,zhang2023interpretable}.
Distinct from traditional RL, RLTF offers only one feedback after a complete trajectory~\citep{arjona2019rudder,zhang2023interpretable}. 
A trajectory $\tau = \{ (s_0, a_0), (s_1, a_1), \ldots, (s_{T-1}, a_{T-1}) \}$ consists of $T$ state-action pairs, with a cumulative reward $R_{\mathrm{traj}}(\tau)$ that is the sum of latent immediate rewards $\sum_{t=0}^{T-1}r(s_t,a_t)$, observable only at the end.
Denoting by $\mathcal T(\pi)$ as the distribution of trajectories induced by $\pi, P, \mu$, the learning objective in RLTF is to maximize the expected trajectory-based reward:
\begin{equation}
    J_\mathrm{traj}(\pi) = \mathbb{E}_{\mathcal T(\pi)} \big[ R_{\mathrm{traj}}(\tau)\big].
\end{equation}

\subsection{Problem Formulation}
\label{sec:Problem Formulation}
We formulate the RLBR problem as an extension of the traditional MDP, the BRMDP, which has a granularity of rewards in between the above two settings. First, we define the notion of {\it bags}, which are sub-pieces of complete trajectories. A trajectory $\tau$ is divided into several neighboring bags, and a bag of size $n_i$, which starts from time $i$, is defined as $B_{i,n_i} = \{(s_i, a_i), \ldots, (s_{i + n_i - 1}, a_{i + n_i - 1})\}, 0\leq i \leq i+n_i-1\leq T-1$. Afterward, we define the BRMDP to navigate the complexities of the aggregated non-Markovian reward signals:

\begin{definition}[BRMDP]
A BRMDP is defined by the tuple $(\mathcal{S}, \mathcal{A}, \mathit{P}, \mathit{R}, \mu)$, where
\begin{itemize}
    \item $\mathcal{S}$ and $\mathcal{A}$ are sets of states and actions.
    \item $\mathit{P}$ is the state transition probability function.
    \item $\mathit{R}$ denotes the bagged reward function, which define the reward over a bag: \( \mathit{R} (B_{i,n_i}) \).
    \item $\mu$ represents the initial state distribution.
\end{itemize}
\end{definition}
In the RLBR framework, a bag $B_{i,n_i}$ metaphorically a unified reward unit for a contiguous sequence of state-action pairs. A trajectory $\tau$ is a composite of a set of bags, denoted as $\mathcal{B}_{\tau}$, which ensures that each trajectory includes at least one reward bag. We further assume a bag partition function defined by the environment: $\mathcal G: \tau\rightarrow \mathcal{B}_\tau$, which is a task-dependent function for generating bags given an input trajectory.
Consequently, the learning objective of the policy is to maximize the accumulated bagged rewards:
\begin{equation}
    J_\mathrm{B}(\pi) = \mathbb{E}_{\mathcal T(\pi)} \bigg[ \sum_{B \in \mathcal{B}_{\tau}} R(B) \bigg| \mathcal G \bigg].
\end{equation}

Notably, if each bag comprises a single instance ($n_i=1, \forall 0 \leq i \leq T-1$), RLBR simplifies to standard RL. Conversely, if a single bag encompasses the entire trajectory ($n_0=T$), RLBR reduces to RLTF. This adaptability highlights the capacity of the RLBR framework to accommodate varying reward structure scenarios.

\begin{figure*}[t]
    \centering
    \includegraphics[width=0.9\textwidth]{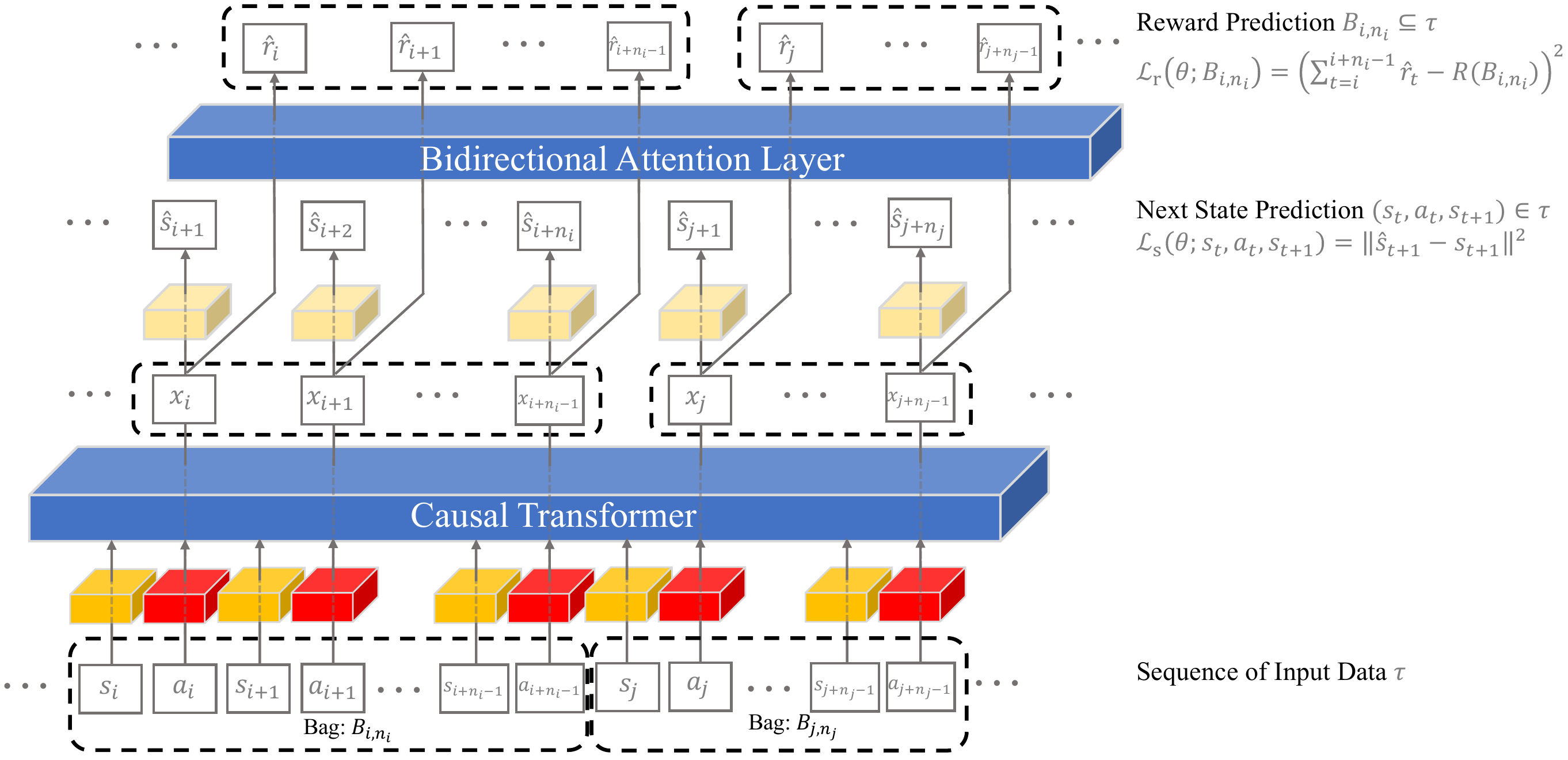}
    \caption{The illustration of the Reward Bag Transformer (RBT) architecture. The Causal Transformer is used for reward representation by processing sequences of input data consisting of state-action pairs. The bidirectional attention layer is used for reward redistribution, utilizing the outputs of the Causal Transformer to predict instance-level rewards. The next state prediction helps the model understand the environment, thereby improving reward prediction.}
    \label{fig:transformer}
    \vspace{-4mm}
\end{figure*}

\section{Reward Redistribution for RLBR}
\label{sec_method}
This section delves into the proposed reward redistribution method for RLBR. We aim for the reward redistribution to meet the following criteria: first, the policies learned from the MDP after reward redistribution should be consistent with those learned from the BRMDP; second, the redistribution process should reflect each instance's contribution based on the sequence information. The subsequent subsections are structured to first theoretically prove the rationality of reward redistribution, followed by proposing the use of a bidirectional attention mechanism for this redistribution, and finally outline a comprehensive algorithm for developing efficient policies within the BRMDP framework.

\subsection{Equivalence of Optimal Policies in the BRMDP and the MDP after Reward Redistribution}
\label{sec:Equivalence of Optimal Policies}
In the BRMDP, agents do not have direct access to the immediate rewards associated with each instance. To address this, we adopt a common approach used in trajectory feedback scenarios, modeling the cumulative sum form as the reward redistribution strategy to convert bagged rewards into instance-level rewards \cite{ren2021rrd,arjona2019rudder,zhang2023interpretable}. Specifically, we define \( \mathit{R} (B_{i,n_i}) = \sum_{t=i}^{i + n_i - 1} \hat{r}(s_t, a_t)\), where \(\hat{r}(s_t, a_t)\) represents the redistributed reward. However, a critical question arises: How can we ensure that the policy learned with redistributed rewards performs well in the original BRMDP? To address this, we provide the following theorem:

\begin{theorem}
\label{theorem_1}
Provide a BRMDP where the bagged reward is regarded as the sum of rewards for state-action pairs of a MDP after reward redistribution contained within the same bag. 
In this context, the set of optimal policies for the BRMDP $\Pi_\mathrm{B}$ aligns with that of the MDP after reward redistribution $\Pi$, implying that $\Pi = \Pi_\mathrm{B}$.
\end{theorem}

The detailed proof of Theorem~\ref{theorem_1} is deferred to the ``Omitted Proof'' in Appendix.
By this theorem, we can see that optimizing a policy on a MDP after reward redistribution is equivalent to optimizing on the original BRMDP.

\subsection{Reward Redistribution based on Bidirectional Attention Mechanism}
Building on Theorem~\ref{theorem_1} that optimal policies in the BRMDP and the MDP after reward redistribution are equivalent, our focus shifts to the crucial process of redistributing rewards within a bag. 
To capture the contextual influence of each instance within the sequence, the Causal Transformer~\cite{vaswani2017attention} is a natural choice as a sequential prediction model. Traditionally, Transformers in RL are used in a unidirectional manner \cite{chen2021decision, janner2021offline, iris2023}, where only previous instances influence the current prediction due to the unobservability of future instances.
However, since our bagged rewards are generally non-Markovian, both preceding and subsequent instances influence the contribution of the current instance to the bagged reward, understanding the relationships among instances within a bag is therefore pivotal. This insight leads us to employ a bidirectional attention mechanism \cite{seo2016bidirectional, vaswani2017attention, DBLP:conf/naacl/DevlinCLT19} to capture these relationships.
This mechanism connects both past and future instances within a bag, enabling a more comprehensive understanding of contextual influences.
By quantitatively evaluating the contribution of each instance, the bidirectional attention mechanism facilitates nuanced and effective reward redistribution, as experimentally demonstrated in Section 5.

\subsection{Reward Bag Transformer}
\label{sec:model architecture}
We introduce the \textit{Reward Bag Transformer} (RBT), a novel reward redistribution method for the BRMDP. The RBT is engineered to comprehend the complex dynamics of the environment through bags and to precisely predict instance-level rewards, facilitating effective reward redistribution.

\subsubsection{Causal Transformer for Reward Representation}
Referring to Fig.~\ref{fig:transformer}, the RBT comprises a Causal Transformer~\citep{vaswani2017attention,radford2018improving}, which maintains the chronological order of state-action pairs \cite{chen2021decision, janner2021offline}. 
For each time step $t$ in a sequence of $M$ time steps, the Causal Transformer, represented as a function $f$, processes the input sequence $\sigma = \{ s_0, a_0, \ldots, s_{M-1}, a_{M-1} \}$, generating the output $\{x_t\}_{t=0}^{M-1}=f(\sigma)$. By aligning the output head $x_t$ with the action token $a_t$, we directly model the consequence of actions, which are pivotal in computing immediate rewards and predicting subsequent states, thereby helping the model better understand environmental dynamics.

\subsubsection{Bidirectional Attention Layer for Reward Redistribution} Once we have obtained the output embeddings \( \{ x_t \}_{t=0}^{M-1} \), for reward prediction, they pass through a bidirectional attention layer to produce \( \{ \hat{r}_t \}_{t=0}^{M-1} \), where $\hat{r}_t \equiv \hat{r}_{\theta}(s_t,a_t)$ with $\theta$ being the RBT parameters. This layer addresses the unidirectional limitation of the Causal Transformer architecture \cite{vaswani2017attention, radford2018improving}, integrating past and future contexts for enhanced reward prediction accuracy. For state prediction, \(x_t\) is input into a state linear decoder, yielding the predicted next state $\hat{s}_{t+1} \equiv \hat{s}_{\theta}(s_t,a_t)$.

The core of the RBT architecture is its bidirectional attention mechanism. For each output embedding $x_t$, we apply three different linear transformations to obtain embeddings for query $\mathbf{q}_t \in \mathbb{R}^{d}$, key $\mathbf{k}_t  \in \mathbb{R}^{d}$, and value $v_t \in \mathbb{R}$, where $d$ is the embedding dimension of the key.
Then the instance-level reward is calculated by
\begin{equation}
    \hat{r}_t = \sum_{\ell=0}^{M-1} \texttt{softmax}(\frac{\{\langle \mathbf{q}_t, \mathbf{k}_{t'} \rangle\}_{t'=0}^{M-1}}{\sqrt{d}})_\ell \cdot v_\ell.
\end{equation}
The rescaling operation is used to prevent extremely small gradients as in~\citet{vaswani2017attention}. 
This mechanism enables the RBT to consider both the immediate and contextual relevance of each instance in a bag when predicting rewards.

\begin{algorithm}[t]
\caption{Policy Optimization with RBT}
\label{alg:policy_learning}
\begin{algorithmic}[1]
    \STATE{Initialize replay buffer $\mathcal{D}$, RBT parameters $\theta$.}
    \FOR{trajectory $\tau$ collected from the environment}
        \STATE{Store trajectory $\tau$ with bag information $\{(B_{i,n_i}, R(B_{i,n_i}))\}_{B_{i,n_i} \in \mathcal{B}_{\tau}}$ in $\mathcal{D}$.} \\
        \STATE{Sample batches from $\mathcal{D}$.}
        \STATE{Estimate bag loss based on Eq.~\eqref{eqn:bag_loss}.}
        \STATE{Update RBT parameters $\theta$ based on the estimated loss.}
        \STATE{Relabel rewards in $\mathcal{D}$ using the updated RBT.}
        \STATE{Optimize policy using the relabeled data by off-the-shelf RL algorithms (such as SAC~\cite{haarnoja2018soft}).}
    \ENDFOR
\end{algorithmic}
\end{algorithm}

\subsection{Learning Objectives}
\label{sec:learning objective}
The learning objectives of the RBT are twofold: \textit{reward prediction} within each reward bag and \textit{state transition forecasting}. These objectives are critical for enabling the model to navigate the complex dynamics of BRMDP environments.

\subsubsection{Reward Prediction}
The RBT is trained to ensure that, for each reward bag, the sum of predicted instance-level rewards matches the total bagged reward. This is vital for maintaining the integrity of the reward structure in the BRMDP framework. The loss function is expressed as
\begin{equation}
\mathcal{L}_\mathrm{r}(\theta; B_{i,n_i}) = \left( \sum_{t=i}^{i+n_i-1} \hat{r}_t - R(B_{i,n_i}) \right)^2.
\end{equation}
This formulation encourages the RBT to learn a nuanced distribution of rewards across states and actions within a bag.
At the same time, it ensures that the sum of redistributed rewards matches the total bagged reward, maintaining consistency as per Theorem~\ref{theorem_1}.

\subsubsection{State Transition Forecasting}
Alongside reward prediction, the RBT is tasked with accurately predicting the next state in the environment given the current state and action. This capability is crucial for understanding the dynamics of the environment. The corresponding loss function is:
\begin{equation}
\mathcal{L}_\mathrm{s}(\theta; s_t, a_t, s_{t+1}) = \| \hat{s}_{t+1} - s_{t+1} \|^2,
\end{equation}
where $\|\cdot\|$ denotes the $\ell^2$-norm. This loss emphasizes the model's understanding of dynamics.

\subsubsection{Composite Loss}
The final learning objective combines the reward and state prediction losses:
\begin{equation}
\begin{aligned}
\label{eqn:bag_loss}
\mathcal{L}_\mathrm{bag}
& (\theta) 
= \underset{\tau \sim \mathcal{D}}{\mathbb{E}} \big[ \mathcal{L}_\mathrm{r}(\theta; B_{i,n_i}) \big| B_{i,n_i} \in \mathcal{B}_\tau \big] \\
& + \beta \underset{\tau \sim \mathcal{D}}{\mathbb{E}} \big[\mathcal{L}_\mathrm{s}(\theta; s_t, a_t, s_{t+1}) \big| (s_t, a_t, s_{t+1}) \in \tau \big],
\end{aligned}
\end{equation}
where the coefficient \( \beta > 0 \) balances the two loss components, and $\mathcal{D}$ denotes the replay buffer.

The RBT's dual predictive capacity is its key advantage, enabling precise reward redistribution to individual instances and forecasting the next state. This leverages environmental dynamics for enhanced reward distribution as experimentally evidenced in Section 5. Integrated with off-the-shelf RL algorithms such as SAC~\cite{haarnoja2018soft}, the RBT can enhance policy learning within the BRMDP framework, as outlined in Algorithm~\ref{alg:policy_learning}.

\section{Experiment}
\label{sec_experiment}

In the following experiment section, we scrutinize the efficacy of our proposed method using benchmark tasks from both the MuJoCo \cite{brockman2016openai} and the DeepMind Control Suite \cite{tassa2018deepmind} environments, focusing on scenarios with bagged rewards.
Initially, we assess the performance of our method to understand its overall effectiveness. Subsequently, we examine whether the proposed RBT reward model accurately predicts rewards. Finally, we evaluate the indispensability of each component of the reward model, questioning if every part is essential.

\subsection{Performance Comparison}
\label{subsec:main}

\subsubsection{Experiment Setting} 
We evaluated our method on benchmark tasks from the MuJoCo locomotion suite (Ant-v2, Hopper-v2, HalfCheetah-v2, and Walker2d-v2) and the DeepMind Control Suite (cheetah-run, quadruped-walk, fish-upright, cartpole-swingup, ball\_in\_cup-catch, and reacher-hard).
Differing from standard environments where rewards are assigned at each step, our approach involved assigning a cumulative reward at the end of each bag while assigning a reward of zero to all other state-action pairs within the bag. The maximum length for each episode was fixed at 1000 steps across all tasks.

\begin{figure*}
  \centering
  \includegraphics[width=1\textwidth]{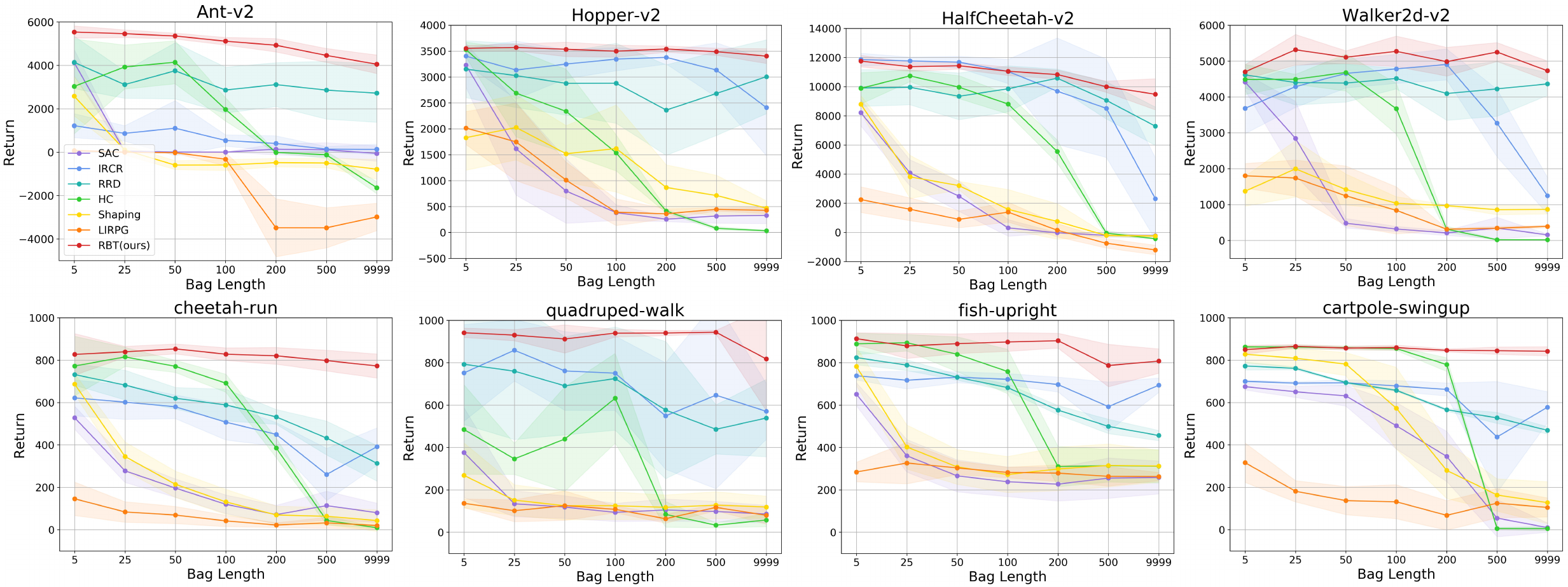}
  \caption{
  Performance comparison across fixed-length reward bag in MuJoCo (top row) and DeepMind Control Suite (bottom row) environments with six different length bag settings (5, 25, 50, 100, 200, and 500) and trajectory feedback (labeled as 9999). The mean and standard deviation are displayed over 6 trials with different random seeds across a total of 1e6 time steps.}
  \label{fig:fixed}
  \vspace{-5mm}
\end{figure*}

\subsubsection{Baselines} In the comparative analysis, our framework was rigorously evaluated against several leading algorithms in the domain of RL with delayed reward:
\begin{itemize}
    \item \textbf{SAC}~\citep{haarnoja2018soft}: It directly utilized the original bagged reward information for policy training using the SAC algorithm.
    \item \textbf{IRCR} \cite{gangwani2020ircr}: It adopted a non-parametric uniform reward redistribution approach. We have adapted IRCR for bagged reward setting.
    \item \textbf{RRD} \cite{ren2021rrd}: It employed a reward model trained with a randomized return decomposition loss. We have adapted RRD for bagged reward setting.
    \item \textbf{HC} \cite{han2022hc}: The HC-decomposition framework was utilized to train the policy using a value function that operates on sequences of data. We employed the code as provided by the original paper.
    \item \textbf{Shaping} \cite{hu2020learning}: A widely used reward shaping method for sparse reward setting, the shaping reward function is adopted from \citet{hu2020learning}.
    \item \textbf{LIRPG} \cite{zheng2018learning}: It learned an intrinsic reward function to complement sparse environmental feedback, training policies to maximize combined extrinsic and intrinsic rewards. We use the same code provided by the paper.
\end{itemize}

\begin{table*}[t]
\caption{Performance comparison across arbitrary reward bag configurations over 6 trials with 1e6 time steps for training, presenting average scores and standard deviations. ``Narrow'' refers to bags with lengths varying arbitrarily from 25 to 200 and narrow intervals between -10 to 10. ``Broad'' denotes the setting with bag lengths varying arbitrarily from 100 to 200 and broad interval variations from -40 to 40. The best and comparable methods based on the paired t-test at the significance level $5\%$ are highlighted in boldface.}
\label{table:arbitrary}
\centering
	\resizebox{1\textwidth}{!}{
\begin{tabular}{c|c|cccccc}
\toprule
Bag Setting  & Environment & SAC & IRCR & RRD & LIRPG & HC & RBT(ours) \\
\midrule
\multirow{8}*{Narrow} & \multirow{2}*{Ant-v2} & 0.87 & 368.69 & 2272.39 & -756.78 & 106.92 & \textbf{5122.50} \\
& & (2.98) & (119.74) & (835.86) & (763.66) & (153.86) & \textbf{(206.44)} \\
& \multirow{2}*{Hopper-v2} & 317.72 & 3353.35 & 2184.41 & 126.13 & 510.66 & \textbf{3499.54} \\
& & (52.17) & (61.97) & (807.71) & (30.18) & (94.49) & \textbf{(76.62)} \\
& \multirow{2}*{HalfCheetah-v2} & 788.45 & \textbf{10853.85} & 9709.62 & 1101.38 & 4027.25 & \textbf{11282.24} \\
& & (1737.57) & \textbf{(573.72)} & (1479.73) & (1248.45) & (441.01) & \textbf{(266.08)} \\
& \multirow{2}*{Walker2d-v2} & 193.07 & 4144.65 & 3536.90 & 123.43 & 309.19 & \textbf{4983.39} \\
& & (48.40) & (673.66) & (546.66) & (50.97) & (171.69) & \textbf{(311.09)} \\

\midrule

\multirow{8}*{Broad}& \multirow{2}*{Ant-v2} & -3.31 & 368.69 & 1323.50 & -1264.08 & 5.97 & \textbf{5167.79} \\
& & (4.15) & (158.46) & (1079.60) & (416.86) & (20.08) & \textbf{(303.83)} \\
& \multirow{2}*{Hopper-v2} & 329.48 & 3296.20 & 1102.38 & 203.01 & 701.84 & \textbf{3499.53} \\
& & (44.21) & (216.35) & (892.12) & (177.80) & (149.44) & \textbf{(94.00)} \\
& \multirow{2}*{HalfCheetah-v2} & 43.96 & 9158.14 & 4199.16 & 924.26 & 4460.80 & \textbf{10837.15} \\
& & (94.32) & (1402.62) & (1476.85) & (1110.97) & (518.94) & \textbf{(254.99)} \\
& \multirow{2}*{Walker2d-v2} & 176.09 & 4179.08 & 330.96 & 194.95 & 447.45 & \textbf{5202.38} \\
& & (49.81) & (937.42) & (79.26) & (98.05) & (155.63) & \textbf{(248.35)} \\
\bottomrule
\end{tabular}}
\end{table*}

While methods like RUDDER \cite{arjona2019rudder} and Align-RUDDER \cite{patil2022align} are known for addressing the problem of trajectory feedback, previous studies \cite{gangwani2020ircr, ren2021rrd, zhang2023interpretable} have shown superior performance using referenced methods. Additionally, since Align-RUDDER relies on successful trajectories for scoring state-action pairs, which is impractical in MuJoCo \cite{patil2022align}, we ultimately excluded both methods from our comparison. 
Sparse reward baselines were included in our comparisons to highlight the distinctions and challenges unique to our bagged reward setting. While both settings provide rewards after a sequence, sparse rewards are Markovian, whereas our bagged rewards are non-Markovian. Therefore, including sparse reward baselines illustrates that directly applying sparse reward methods to bagged rewards is possible but not effective. For more discussion on sparse rewards, see ``Discussion on Sparse Rewards'' in Appendix.
Besides, detailed descriptions of the model parameters and hyper-parameters used during training are provided in ``Experiment Settings and Implementation Details'', with more experimental results available in ``Additional Experimental Results'' in Appendix. 

\subsubsection{Evaluation Metric}
We report the average accumulative reward across 6 seeds with random initialization to demonstrate the performance of evaluated methods. Higher accumulative reward in evaluation indicates better performance.

\subsubsection{Experimental Results}
In the \textbf{fixed-length} reward bag experiment (see Fig.~\ref{fig:fixed}), we conducted experiments with six bag lengths (5, 25, 50, 100, 200, and 500) and trajectory feedback (labeled as 9999) across each environment, where bagged reward is made by accumulating immediate rewards within bags. This aimed to illustrate the influence of varying bag lengths, providing insight into how bag size affected the performance of the learning algorithm. 
The SAC method, using bagged rewards directly from the environment, suffers from a lack of guidance in agent training due to missing reward information. This issue worsens with longer bag lengths, indicating that increased reward sparsity leads to less effective policy optimization. The IRCR and RRD methods, treating rewards uniformly within a reward bag, outperform SAC, suggesting benefits from even approximate guidance. However, notable variance in their results indicates potential consistency and reliability issues. 
The HC method excels only with shorter bag lengths, suggesting that this value function modification method struggles to utilize information from longer sequences. 
The Shaping and LIRPG methods exhibit subdued performance across tasks, as they are designed to solve sparse reward problems with Markovian rewards, which do not align with the reward bag setting.
The proposed RBT method consistently outperforms the other approaches across all the environments and bag lengths, showing that it is not only well-suited for environments with short reward bags but also capable of handling large reward bag scenarios. This demonstrated the capability of RBT to learn from the sequence of instances and, by integrating bagged reward information, accurately allocate rewards to instances, thereby guiding better policy training.

To validate the effectiveness of our approach under more complex conditions, we designed an experiment with \textbf{arbitrary-length} reward bags, allowing for overlaps or gaps between reward bags. This setup simulated more realistic scenarios and tested the robustness of our method. The results, detailed in Table~\ref{table:arbitrary}, confirm the superior performance of the proposed RBT method in these complex reward settings. These findings suggest RBT's potential for broader applications, demonstrating its versatility and robustness in handling intricate reward dynamics.

We also conducted experiments in \textbf{sparse reward environments}, where tasks are rewarded with 1 at the end of the trajectory if completed and 0 otherwise. In this setting, the reward does not accumulate as in our redistribution model. As shown in Table~\ref{table:sparse}, RLTF methods do not perform well in sparse reward environments. Our proposed method not only outperforms RLTF baselines but also exceeds the performance of sparse reward baselines. 

\begin{table*}[t]
\caption{Performance comparison on sparse reward environments over 6 trials with 1e6 time steps for training,
presenting average scores and standard deviations. Rewards are given with 1 at the end of the trajectory if completed and 0 otherwise. The best and comparable methods based on the paired t-test at the significance level $5\%$ are
highlighted in boldface.}
\label{table:sparse}
\centering
\fontsize{9}{9}\selectfont{
\begin{tabular}{c|ccccccc}
\toprule
Environment & SAC & IRCR & RRD & HC & Shaping & LIRPG & RBT(ours) \\
\midrule
\multirow{2}*{ball\_in\_cup-catch} & 646.90 & 773.19 & 666.18 & 76.19 & 847.88 & 125.50 & \textbf{972.19} \\
 & (231.15) & (73.06) & (33.02) & (67.25) & (178.01) & (21.25) & \textbf{(5.50)} \\

\midrule

\multirow{2}*{reacher-hard} & 531.88 & 584.76 & 347.16 & 7.57 & 664.49 & 8.74 & \textbf{735.67} \\
 & (127.94) & (269.32) & (106.56) & (3.50) & (170.50) & (2.36) & \textbf{(144.90)} \\

\bottomrule
\end{tabular}}
\end{table*}

\subsection{Case Study}

\begin{figure*}[t]
  \centering
  \includegraphics[width=1\textwidth]{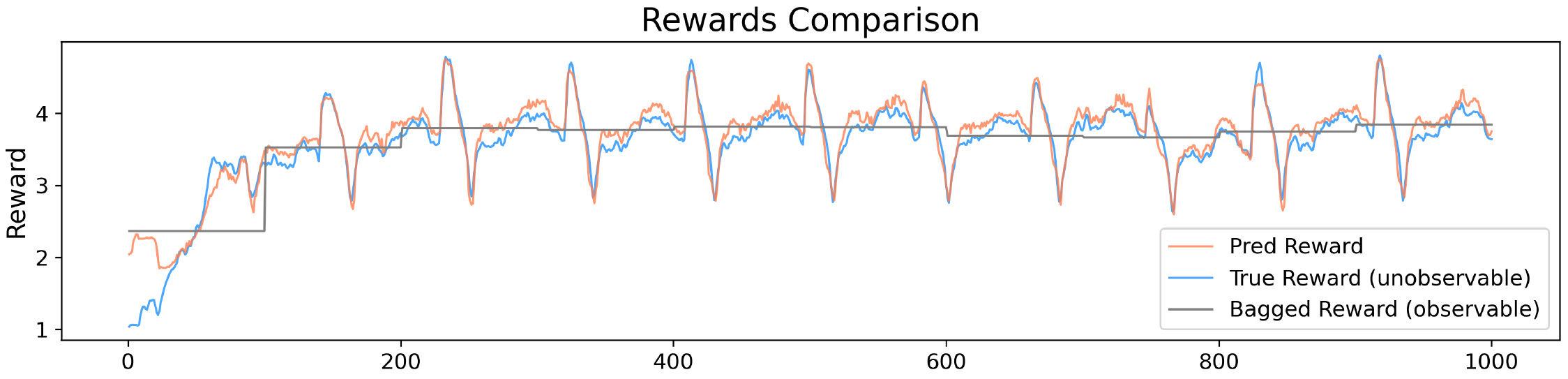}
  \caption{Comparison of predicted rewards with true rewards and aggregated bagged rewards.}
  \label{fig:reward_comparison}
\end{figure*}

The previous experimental results showcase the superiority of RBT over baselines. This led to an intriguing inquiry: Is the RBT reward model proficient in accurately redistributing rewards? 
To investigate this question, we performed an experiment focused on reward comparison, utilizing a trajectory generated by an agent trained in the Hopper-v2 environment with a bag length of 100.
As shown in Fig.~\ref{fig:reward_comparison}, which spans 1000 steps, RBT-predicted rewards, unobservable true rewards, and observable bagged rewards (presented in a uniform format for better visualization) are compared.
The figure indicates that the rewards predicted by the RBT closely match the trends of the true rewards. This observation suggest that the RBT is effective at reconstructing true rewards from bagged rewards, despite the coarse nature of the environmental reward signals.

\begin{figure*}[t]
  \centering
  \includegraphics[width=1\textwidth]{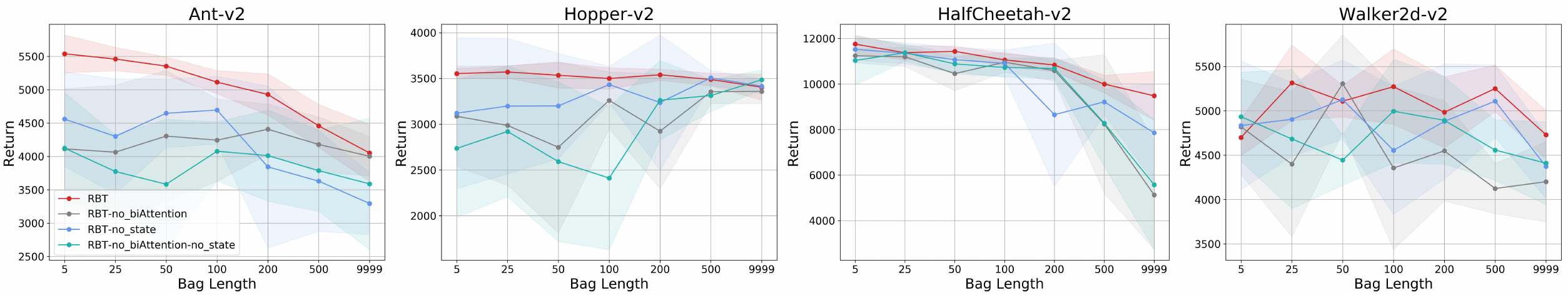}
  \caption{Ablation study of reward model components across various environments. The chart presents mean and standard deviation of rewards over 6 trials with 1e6 timesteps, showcasing the efficacy of the full proposed method relative to its variants without certain features.}
  \label{fig:ablation}
\end{figure*}

\subsection{Ablation Study}
\label{app:sec_ablation_study}
We conducted comparisons to examine the role of RBT's modules. As shown in Fig.~\ref{fig:ablation}, the full RBT model consistently outperforms its variants, indicating a synergistic effect when all components are used together. 
Performance drops significantly when the bidirectional attention mechanism is removed, especially in complex environments like Ant-v2 and HalfCheetah-v2, suggesting its critical role in accurate reward prediction. Additionally, we can observe that removing the next state prediction component weakens the reward model's understanding of environmental dynamics, reducing reward prediction accuracy and hindering policy learning. The greatest performance decline occurs when both the next state prediction and bidirectional self-attention mechanism are absent, underscoring their individual and combined importance in building a robust reward model.

\section{Conclusion}
\label{sec_conclusin}
In this paper, we introduced a novel learning framework called Reinforcement Learning from Bagged Rewards (RLBR). 
To address the challenges posed by this framework, we established theoretical connections between Bagged Reward MDPs (BRMDPs) and MDPs after reward redistribution, providing a solid foundation for our approach. Based on this theory, we proposed a reward model, the Reward Bag Transformer (RBT), designed to efficiently redistribute rewards by interpreting contextual information within bags and understanding environmental dynamics.
The efficacy of RBT was demonstrated through extensive experiments, where it consistently outperformed existing methods across various reward bag scenarios. Additionally, our case studies highlighted RBT's ability to effectively reallocate rewards while maintaining fidelity to the underlying true reward structure.

\bibliography{main}

\begin{thebibliography}{46}
\providecommand{\natexlab}[1]{#1}
\providecommand{\url}[1]{\texttt{#1}}
\expandafter\ifx\csname urlstyle\endcsname\relax
  \providecommand{\doi}[1]{doi: #1}\else
  \providecommand{\doi}{doi: \begingroup \urlstyle{rm}\Url}\fi

\bibitem[Arjona-Medina et~al.(2019)Arjona-Medina, Gillhofer, Widrich, Unterthiner, Brandstetter, and Hochreiter]{arjona2019rudder}
Jose~A Arjona-Medina, Michael Gillhofer, Michael Widrich, Thomas Unterthiner, Johannes Brandstetter, and Sepp Hochreiter.
\newblock Rudder: Return decomposition for delayed rewards.
\newblock In \emph{The Thirty-second Annual Conference on Advances in Neural Information Processing Systems}, 2019.

\bibitem[Brockman et~al.(2016)Brockman, Cheung, Pettersson, Schneider, Schulman, Tang, and Zaremba]{brockman2016openai}
Greg Brockman, Vicki Cheung, Ludwig Pettersson, Jonas Schneider, John Schulman, Jie Tang, and Wojciech Zaremba.
\newblock Openai gym, 2016.

\bibitem[Burda et~al.(2019)Burda, Edwards, Storkey, and Klimov]{burda2019exploration}
Yuri Burda, Harrison Edwards, Amos Storkey, and Oleg Klimov.
\newblock Exploration by random network distillation.
\newblock In \emph{The Seventh International Conference on Learning Representations}, 2019.

\bibitem[Chen et~al.(2021)Chen, Lu, Rajeswaran, Lee, Grover, Laskin, Abbeel, Srinivas, and Mordatch]{chen2021decision}
Lili Chen, Kevin Lu, Aravind Rajeswaran, Kimin Lee, Aditya Grover, Misha Laskin, Pieter Abbeel, Aravind Srinivas, and Igor Mordatch.
\newblock Decision transformer: Reinforcement learning via sequence modeling.
\newblock In \emph{The Thirty-fourth Annual Conference on Advances in Neural Information Processing Systems}, 2021.

\bibitem[Devlin et~al.(2019)Devlin, Chang, Lee, and Toutanova]{DBLP:conf/naacl/DevlinCLT19}
Jacob Devlin, Ming{-}Wei Chang, Kenton Lee, and Kristina Toutanova.
\newblock {BERT:} pre-training of deep bidirectional transformers for language understanding.
\newblock In Jill Burstein, Christy Doran, and Thamar Solorio, editors, \emph{Proceedings of the 2019 Conference of the North American Chapter of the Association for Computational Linguistics: Human Language Technologies, {NAACL-HLT} 2019, Minneapolis, MN, USA, June 2-7, 2019, Volume 1 (Long and Short Papers)}, pages 4171--4186. Association for Computational Linguistics, 2019.

\bibitem[Early et~al.(2022)Early, Bewley, Evers, and Ramchurn]{early2022non}
Joseph Early, Tom Bewley, Christine Evers, and Sarvapali Ramchurn.
\newblock Non-markovian reward modelling from trajectory labels via interpretable multiple instance learning.
\newblock In \emph{The Thirty-fifth Annual Conference on Advances in Neural Information Processing Systems}, 2022.

\bibitem[Florensa et~al.(2018)Florensa, Held, Geng, and Abbeel]{florensa2018automatic}
Carlos Florensa, David Held, Xinyang Geng, and Pieter Abbeel.
\newblock Automatic goal generation for reinforcement learning agents.
\newblock In \emph{The Thirty-fifth International Conference on Machine Learning}. PMLR, 2018.

\bibitem[Frostig et~al.(2018)Frostig, Johnson, and Leary]{frostig2018compiling}
Roy Frostig, Matthew~James Johnson, and Chris Leary.
\newblock Compiling machine learning programs via high-level tracing.
\newblock \emph{Systems for Machine Learning}, 4\penalty0 (9), 2018.

\bibitem[Gangwani et~al.(2020)Gangwani, Zhou, and Peng]{gangwani2020ircr}
Tanmay Gangwani, Yuan Zhou, and Jian Peng.
\newblock Learning guidance rewards with trajectory-space smoothing.
\newblock In \emph{The Thirty-third Annual Conference on Advances in Neural Information Processing Systems}, 2020.

\bibitem[Gaon and Brafman(2020)]{gaon2020reinforcement}
Maor Gaon and Ronen Brafman.
\newblock Reinforcement learning with non-markovian rewards.
\newblock In \emph{The Thirty-Fourth AAAI Conference on Artificial Intelligence}, volume~34, 2020.

\bibitem[Haarnoja et~al.(2018)Haarnoja, Zhou, Abbeel, and Levine]{haarnoja2018soft}
Tuomas Haarnoja, Aurick Zhou, Pieter Abbeel, and Sergey Levine.
\newblock Soft actor-critic: Off-policy maximum entropy deep reinforcement learning with a stochastic actor.
\newblock In \emph{The Thirty-fifth International Conference on Machine Learning}. PMLR, 2018.

\bibitem[Han et~al.(2022)Han, Ren, Wu, Zhou, and Peng]{han2022hc}
Beining Han, Zhizhou Ren, Zuofan Wu, Yuan Zhou, and Jian Peng.
\newblock Off-policy reinforcement learning with delayed rewards.
\newblock In \emph{The Thirty-ninth International Conference on Machine Learning}. PMLR, 2022.

\bibitem[Hu et~al.(2020)Hu, Wang, Jia, Wang, Chen, Hao, Wu, and Fan]{hu2020learning}
Yujing Hu, Weixun Wang, Hangtian Jia, Yixiang Wang, Yingfeng Chen, Jianye Hao, Feng Wu, and Changjie Fan.
\newblock Learning to utilize shaping rewards: A new approach of reward shaping.
\newblock In \emph{The Thirty-third Annual Conference on Advances in Neural Information Processing Systems}, 2020.

\bibitem[Janner et~al.(2021)Janner, Li, and Levine]{janner2021offline}
Michael Janner, Qiyang Li, and Sergey Levine.
\newblock Offline reinforcement learning as one big sequence modeling problem.
\newblock In \emph{The Thirty-fourth Annual Conference on Advances in Neural Information Processing Systems}, 2021.

\bibitem[Ke et~al.(2018)Ke, ALIAS PARTH~GOYAL, Bilaniuk, Binas, Mozer, Pal, and Bengio]{ke2018sparse}
Nan~Rosemary Ke, Anirudh~Goyal ALIAS PARTH~GOYAL, Olexa Bilaniuk, Jonathan Binas, Michael~C Mozer, Chris Pal, and Yoshua Bengio.
\newblock Sparse attentive backtracking: Temporal credit assignment through reminding.
\newblock In \emph{The Thirty-first Annual Conference on Advances in Neural Information Processing Systems}, 2018.

\bibitem[Kiran et~al.(2021)Kiran, Sobh, Talpaert, Mannion, Al~Sallab, Yogamani, and P{\'e}rez]{kiran2021deep}
B~Ravi Kiran, Ibrahim Sobh, Victor Talpaert, Patrick Mannion, Ahmad~A Al~Sallab, Senthil Yogamani, and Patrick P{\'e}rez.
\newblock Deep reinforcement learning for autonomous driving: A survey.
\newblock \emph{IEEE Transactions on Intelligent Transportation Systems}, 23\penalty0 (6):\penalty0 4909--4926, 2021.

\bibitem[Kostrikov(2021)]{jaxrl}
Ilya Kostrikov.
\newblock {JAXRL: Implementations of Reinforcement Learning algorithms in JAX}, 10 2021.
\newblock URL \url{https://github.com/ikostrikov/jaxrl}.

\bibitem[Kwon et~al.(2023)Kwon, Xie, Bullard, and Sadigh]{kwonreward}
Minae Kwon, Sang~Michael Xie, Kalesha Bullard, and Dorsa Sadigh.
\newblock Reward design with language models.
\newblock In \emph{The Eleventh International Conference on Learning Representations}, 2023.

\bibitem[Lee et~al.(2023)Lee, Phatale, Mansoor, Lu, Mesnard, Bishop, Carbune, and Rastogi]{lee2023rlaif}
Harrison Lee, Samrat Phatale, Hassan Mansoor, Kellie Lu, Thomas Mesnard, Colton Bishop, Victor Carbune, and Abhinav Rastogi.
\newblock Rlaif: Scaling reinforcement learning from human feedback with ai feedback, 2023.

\bibitem[Liu et~al.(2019)Liu, Luo, Zhong, Chen, Liu, and Peng]{liu2019sequence}
Yang Liu, Yunan Luo, Yuanyi Zhong, Xi~Chen, Qiang Liu, and Jian Peng.
\newblock Sequence modeling of temporal credit assignment for episodic reinforcement learning, 2019.

\bibitem[Loshchilov and Hutter(2018)]{loshchilov2018decoupled}
Ilya Loshchilov and Frank Hutter.
\newblock Decoupled weight decay regularization.
\newblock In \emph{The Sixth International Conference on Learning Representations}, 2018.

\bibitem[Micheli et~al.(2023)Micheli, Alonso, and Fleuret]{iris2023}
Vincent Micheli, Eloi Alonso, and Fran{\c{c}}ois Fleuret.
\newblock Transformers are sample-efficient world models.
\newblock In \emph{The Eleventh International Conference on Learning Representations}, 2023.

\bibitem[Ng et~al.(1999)Ng, Harada, and Russell]{ng1999policy}
Andrew~Y Ng, Daishi Harada, and Stuart Russell.
\newblock Policy invariance under reward transformations: Theory and application to reward shaping.
\newblock In \emph{The Sixteenth International Conference on Machine Learning}. PMLR, 1999.

\bibitem[Pathak et~al.(2017)Pathak, Agrawal, Efros, and Darrell]{pathak2017curiosity}
Deepak Pathak, Pulkit Agrawal, Alexei~A Efros, and Trevor Darrell.
\newblock Curiosity-driven exploration by self-supervised prediction.
\newblock In \emph{The Thirty-fourth International Conference on Machine Learning}. PMLR, 2017.

\bibitem[Patil et~al.(2022)Patil, Hofmarcher, Dinu, Dorfer, Blies, Brandstetter, Arjona-Medina, and Hochreiter]{patil2022align}
Vihang Patil, Markus Hofmarcher, Marius-Constantin Dinu, Matthias Dorfer, Patrick~M Blies, Johannes Brandstetter, Jos{\'e} Arjona-Medina, and Sepp Hochreiter.
\newblock Align-rudder: Learning from few demonstrations by reward redistribution.
\newblock In \emph{The Thirty-ninth International Conference on Machine Learning}. PMLR, 2022.

\bibitem[Radford et~al.(2018)Radford, Narasimhan, Salimans, Sutskever, et~al.]{radford2018improving}
Alec Radford, Karthik Narasimhan, Tim Salimans, Ilya Sutskever, et~al.
\newblock Improving language understanding by generative pre-training, 2018.

\bibitem[Ren et~al.(2021)Ren, Guo, Zhou, and Peng]{ren2021rrd}
Zhizhou Ren, Ruihan Guo, Yuan Zhou, and Jian Peng.
\newblock Learning long-term reward redistribution via randomized return decomposition.
\newblock In \emph{The Ninth International Conference on Learning Representations}, 2021.

\bibitem[Riedmiller et~al.(2018)Riedmiller, Hafner, Lampe, Neunert, Degrave, Wiele, Mnih, Heess, and Springenberg]{riedmiller2018learning}
Martin Riedmiller, Roland Hafner, Thomas Lampe, Michael Neunert, Jonas Degrave, Tom Wiele, Vlad Mnih, Nicolas Heess, and Jost~Tobias Springenberg.
\newblock Learning by playing solving sparse reward tasks from scratch.
\newblock In \emph{The Thirty-fifth International Conference on Machine Learning}. PMLR, 2018.

\bibitem[Sekar et~al.(2020)Sekar, Rybkin, Daniilidis, Abbeel, Hafner, and Pathak]{sekar2020planning}
Ramanan Sekar, Oleh Rybkin, Kostas Daniilidis, Pieter Abbeel, Danijar Hafner, and Deepak Pathak.
\newblock Planning to explore via self-supervised world models.
\newblock In \emph{The Thirty-seventh International Conference on Machine Learning}. PMLR, 2020.

\bibitem[Seo et~al.(2016)Seo, Kembhavi, Farhadi, and Hajishirzi]{seo2016bidirectional}
Minjoon Seo, Aniruddha Kembhavi, Ali Farhadi, and Hannaneh Hajishirzi.
\newblock Bidirectional attention flow for machine comprehension.
\newblock In \emph{The Fourth International Conference on Learning Representations}, 2016.

\bibitem[Silver et~al.(2016)Silver, Huang, Maddison, Guez, Sifre, Van Den~Driessche, Schrittwieser, Antonoglou, Panneershelvam, Lanctot, et~al.]{silver2016mastering}
David Silver, Aja Huang, Chris~J Maddison, Arthur Guez, Laurent Sifre, George Van Den~Driessche, Julian Schrittwieser, Ioannis Antonoglou, Veda Panneershelvam, Marc Lanctot, et~al.
\newblock Mastering the game of go with deep neural networks and tree search.
\newblock \emph{nature}, 529\penalty0 (7587):\penalty0 484--489, 2016.

\bibitem[Sorg et~al.(2010)Sorg, Lewis, and Singh]{sorg2010reward}
Jonathan Sorg, Richard~L Lewis, and Satinder Singh.
\newblock Reward design via online gradient ascent.
\newblock In \emph{The Twenty-fourth Annual Conference on Advances in Neural Information Processing Systems}, 2010.

\bibitem[Sutton and Barto(1998)]{712192}
R.S. Sutton and A.G. Barto.
\newblock Reinforcement learning: An introduction.
\newblock \emph{IEEE Transactions on Neural Networks}, 9\penalty0 (5):\penalty0 1054--1054, 1998.

\bibitem[Tambwekar et~al.(2019)Tambwekar, Dhuliawala, Martin, Mehta, Harrison, and Riedl]{tambwekar2019controllable}
Pradyumna Tambwekar, Murtaza Dhuliawala, Lara~J. Martin, Animesh Mehta, Brent Harrison, and Mark~O. Riedl.
\newblock Controllable neural story plot generation via reward shaping.
\newblock In \emph{Proceedings of the Twenty-Eighth International Joint Conference on Artificial Intelligence}. International Joint Conferences on Artificial Intelligence Organization, 2019.

\bibitem[Tassa et~al.(2018)Tassa, Doron, Muldal, Erez, Li, Casas, Budden, Abdolmaleki, Merel, Lefrancq, et~al.]{tassa2018deepmind}
Yuval Tassa, Yotam Doron, Alistair Muldal, Tom Erez, Yazhe Li, Diego de~Las Casas, David Budden, Abbas Abdolmaleki, Josh Merel, Andrew Lefrancq, et~al.
\newblock Deepmind control suite, 2018.

\bibitem[Tessler et~al.(2019)Tessler, Mankowitz, and Mannor]{tesslerreward}
Chen Tessler, Daniel~J Mankowitz, and Shie Mannor.
\newblock Reward constrained policy optimization.
\newblock In \emph{The Seventh International Conference on Learning Representations}, 2019.

\bibitem[Vaswani et~al.(2017)Vaswani, Shazeer, Parmar, Uszkoreit, Jones, Gomez, Kaiser, and Polosukhin]{vaswani2017attention}
Ashish Vaswani, Noam Shazeer, Niki Parmar, Jakob Uszkoreit, Llion Jones, Aidan~N Gomez, {\L}ukasz Kaiser, and Illia Polosukhin.
\newblock Attention is all you need.
\newblock In \emph{The Thirtieth Annual Conference on Advances in Neural Information Processing Systems}, 2017.

\bibitem[Vinyals et~al.(2019)Vinyals, Babuschkin, Czarnecki, Mathieu, Dudzik, Chung, Choi, Powell, Ewalds, Georgiev, et~al.]{vinyals2019grandmaster}
Oriol Vinyals, Igor Babuschkin, Wojciech~M Czarnecki, Michaël Mathieu, Andrew Dudzik, Junyoung Chung, David~H Choi, Richard Powell, Timo Ewalds, Petko Georgiev, et~al.
\newblock Grandmaster level in starcraft ii using multi-agent reinforcement learning.
\newblock \emph{Nature}, 2019.

\bibitem[Watkins(1989)]{watkins1989learning}
Christopher John Cornish~Hellaby Watkins.
\newblock \emph{Learning from delayed rewards}.
\newblock King's College, Cambridge United Kingdom, 1989.

\bibitem[Widrich et~al.(2021)Widrich, Hofmarcher, Patil, Bitto-Nemling, and Hochreiter]{widrich2021modern}
Michael Widrich, Markus Hofmarcher, Vihang~Prakash Patil, Angela Bitto-Nemling, and Sepp Hochreiter.
\newblock Modern hopfield networks for return decomposition for delayed rewards.
\newblock In \emph{Deep RL Workshop NeurIPS 2021}, 2021.

\bibitem[Wurman et~al.(2022)Wurman, Barrett, Kawamoto, MacGlashan, Subramanian, Walsh, Capobianco, Devlic, Eckert, Fuchs, et~al.]{wurman2022outracing}
Peter~R Wurman, Samuel Barrett, Kenta Kawamoto, James MacGlashan, Kaushik Subramanian, Thomas~J Walsh, Roberto Capobianco, Alisa Devlic, Franziska Eckert, Florian Fuchs, et~al.
\newblock Outracing champion gran turismo drivers with deep reinforcement learning.
\newblock \emph{Nature}, 602\penalty0 (7896):\penalty0 223--228, 2022.

\bibitem[Yang et~al.(2020)Yang, Liu, Zhong, and Walid]{yang2020deep}
Hongyang Yang, Xiao-Yang Liu, Shan Zhong, and Anwar Walid.
\newblock Deep reinforcement learning for automated stock trading: An ensemble strategy.
\newblock In \emph{Proceedings of the first ACM international conference on AI in finance}, 2020.

\bibitem[Yu et~al.(2021)Yu, Liu, Nemati, and Yin]{yu2021reinforcement}
Chao Yu, Jiming Liu, Shamim Nemati, and Guosheng Yin.
\newblock Reinforcement learning in healthcare: A survey.
\newblock \emph{ACM Computing Surveys (CSUR)}, 55\penalty0 (1):\penalty0 1--36, 2021.

\bibitem[Zhang et~al.(2023)Zhang, Du, Huang, Wang, Wang, Fang, and Pechenizkiy]{zhang2023interpretable}
Yudi Zhang, Yali Du, Biwei Huang, Ziyan Wang, Jun Wang, Meng Fang, and Mykola Pechenizkiy.
\newblock Interpretable reward redistribution in reinforcement learning: A causal approach.
\newblock In \emph{The Thirty-sixth Annual Conference on Advances in Neural Information Processing Systems}, 2023.

\bibitem[Zheng et~al.(2021)Zheng, Chen, Wang, He, Hu, Chen, Fan, Gao, and Zhang]{zheng2021episodic}
Lulu Zheng, Jiarui Chen, Jianhao Wang, Jiamin He, Yujing Hu, Yingfeng Chen, Changjie Fan, Yang Gao, and Chongjie Zhang.
\newblock Episodic multi-agent reinforcement learning with curiosity-driven exploration.
\newblock In \emph{The Thirty-fourth Annual Conference on Advances in Neural Information Processing Systems}, 2021.

\bibitem[Zheng et~al.(2018)Zheng, Oh, and Singh]{zheng2018learning}
Zeyu Zheng, Junhyuk Oh, and Satinder Singh.
\newblock On learning intrinsic rewards for policy gradient methods.
\newblock In \emph{The Thirty-first Annual Conference on Advances in Neural Information Processing Systems}, 2018.

\end{thebibliography}

\newpage
\appendix
\onecolumn 

\section{Broader Impact}
\label{app:Broader Impact}
In this work, we introduce the problem of Reinforcement Learning from Bagged Reward (RLBR), and propose a Transformer-based reward model that uses bidirectional attention to interpret context within a bag and accurately allocate rewards to each instance.
On the one hand, we recognize that these techniques could raise some potential issues. As collecting bagged rewards is much more convenient and natural than gathering instance-level rewards, this could lead to some risks of abusing unauthorized data. On the other hand, we believe that developing these techniques is still necessary for the importance of solving reinforcement learning tasks with bag-level feedback. Furthermore, there are many techniques for preserving data privacy, which can be compatible with our approach to avoid such problems.

\section{Omitted Proof}
\label{app:proof}
\setcounter{theorem}{0}
\begin{theorem}
Provide a BRMDP where the bagged reward is regarded as the sum of rewards for state-action pairs of a MDP after reward redistribution contained within the same bag. 
In this context, the set of optimal policies for the BRMDP $\Pi_\mathrm{B}$ aligns with that of the MDP after reward redistribution $\Pi$, implying that $\Pi = \Pi_\mathrm{B}$.
\end{theorem}

\begin{proof}
For a given bag \( \mathit{{B}_{i,n_i}} \), if the bagged reward $R(\mathit{{B}_{i,n_i}})$ is regarded as the sum of the individual rewards of the state-action pairs contained within it, we can get:
\begin{equation*}
    R(\mathit{{B}_{i,n_i}}) = \sum_{t=i}^{i+n_i-1} \hat{r}(s_t, a_t).
\end{equation*}

Over a complete trajectory $\tau$, the cumulative reward in BRMDP can be expressed as the sum of the rewards from all the bags along the trajectory:
\begin{equation*}
\begin{aligned}
    \sum_{B \in \mathcal{B}_{\tau}} R(B)
    & = \sum_{i \in \mathcal{I}_{\tau}} R(\mathit{{B}_{i,n_i}}) \\
    & = \sum_{i \in \mathcal{I}_{\tau}} \bigg( \sum_{t=i}^{i+n_i-1} \hat{r}(s_t, a_t) \bigg)
    = \sum_{t=0}^{T-1} \hat{r}(s_t, a_t),
\end{aligned}
\end{equation*}
where $\mathcal{I}_{\tau}$ denotes the set of initial timestep indices for $B \in \mathcal{B}_{\tau}$.

The policy optimization objective in BRMDP, $J_\mathrm{B}(\pi)$, aims to maximize the expected sum of bagged rewards along the trajectory from $t=0$, which is equivalent to maximizing the cumulative reward in redistributed reward MDP, $J(\pi)$:
\begin{equation*}
\begin{aligned}
    J_\mathrm{B}(\pi) 
    & = \mathbb{E}_{\mathcal T(\pi)} \bigg[ \sum_{B \in \mathcal{B}_{\tau}} R(B) \bigg| \mathcal G \bigg] \\
    & = \mathbb{E}_{\pi, P, \mu} \bigg[ \sum_{t=0}^{T-1} \hat{r}(s_t, a_{t}) \bigg]
    = J(\pi).
\end{aligned}
\end{equation*}

Given that the expected cumulative rewards for any policy in the original BRMDP and redistributed reward MDP frameworks are equivalent, under the condition of infinite exploration or exhaustive sampling within the state-action space, the sets of optimal policies for each framework also coincide, implying that $\Pi = \Pi_\mathrm{B}$.
\end{proof}

\begin{table*}[ht]
\centering
\caption{Hyper-parameters of RBT.}
\label{table:hyperparameters_RBT}
\vskip 0.15in
\begin{tabular}{cc}
\hline
Hyper-parameter & Value \\
\hline
Number of Causal Transformer layers & 3 \\
Number of bidirectional attention layers & 1 \\
Number of attention heads & 4 \\
Embedding dimension & 256 \\
Batch size & 64 \\
Dropout rate & 0.1 \\
Learning rate & 0.0001 \\
Optimizer & AdamW \cite{loshchilov2018decoupled} \\
Weight decay & 0.0001 \\
Warmup steps & 100 \\
Total gradient steps & 10000 \\
\hline
\end{tabular}
\end{table*}

\section{Experiment Settings and Implementation Details}
\label{app:experiment}

\paragraph{Benchmarks with Bagged Rewards.}
We introduced a novel problem setting in the suite of MuJoCo and DeepMind Control Suite locomotion benchmark tasks, termed as bagged rewards.
Our simulations ran on the OpenAI Gym platform \cite{brockman2016openai} and the DeepMind Control Suite \cite{tassa2018deepmind}, featuring tasks that stretched over long horizons with a set maximum trajectory length of $T=1000$. We used MuJoCo version 2.0 for our simulations, which is available at \url{http://www.mujoco.org/}. MuJoCo is licensed under a commercial license, and we have adhered to its terms of service and licensing agreements as stated on the official website. The DeepMind Control Suite is available under an Apache License 2.0, and we have complied with its terms of use.

Reward bag experiments of different bag sizes (5, 25, 50, 100, 200, and 500) and trajectory feedback were set up to verify the effectiveness of the method.
To evaluate the efficacy of proposed method, commonly used trajectory feedback algorithms were adapted to fit the bagged reward setting as baselines. In these experiments, each reward bag was treated as an individual trajectory, and these modified algorithms were applied accordingly.
Additionally, experiments using standard trajectory feedback were conducted to provide a comparative baseline within the unique setting. The total episodic feedback was computed at the end of the trajectory and was the sum of the per-step rewards the agent had collected throughout the episode. This experiment setting was the same as some previous works for learning from trajectory feedback \cite{gangwani2020ircr, ren2021rrd}.

For the sparse reward experiments, the ball\_in\_cup-catch task provides a sparse reward of 1 when the ball is in the cup and 0 otherwise; in the reacher-hard task, the reward is 1 when the end effector penetrates the target sphere~\cite{tassa2018deepmind}. Therefore, we treated the sparse reward as trajectory feedback and conducted experiments where the entire trajectory was considered as a single reward bag in these environments. The bagged reward was identical to the sparse reward provided by the environment.

\paragraph{Implementation Details and Hyper-parameter Configuration.}

In our experiments, the policy optimization module was implemented based on soft actor-critic (SAC) \cite{haarnoja2018soft}. We evaluated the performance of our proposed methods with the same configuration of hyper-parameters in all environments. The back-end SAC followed the JaxRL implementation \cite{jaxrl}, which is available under the MIT License.

The RBT reward model was developed based on the GPT implementation in JAX \cite{frostig2018compiling}, which is available under the Apache License 2.0. Our experiments utilized the Causal Transformer with three layers and four self-attention heads, followed by a bidirectional self-attention layer with one self-attention head. For detailed hyper-parameter settings of the RBT, please refer to Table~\ref{table:hyperparameters_RBT}.

For the baseline methods, the IRCR \cite{gangwani2020ircr} method was implemented based on the descriptions provided in the original paper. The RRD \cite{ren2021rrd} and LIRPG \cite{zheng2018learning} methods are both licensed under the MIT License. The code of HC \cite{han2022hc} is available in the supplementary material at \url{https://openreview.net/forum?id=nsjkNB2oKsQ}. 

\begin{table*}[t]
\caption{Performance comparison across reward bag with various-length configurations over 3 trials. In this table, ``Short'' refers to bags with lengths varying from 25 to 200, and ``Long'' denotes the setting with bag lengths from 100 to 500. The best and comparable methods based on the paired t-test at the significance level $5\%$ are highlighted in boldface.}
\label{table:change_len}
\vskip 0.15in
\centering
\resizebox{1\textwidth}{!}{
\begin{tabular}{c|c|cccccc}
\toprule
Bag Setting  & Environment & SAC & IRCR & RRD & LIRPG & HC & RBT(ours) \\
\midrule
\multirow{8}*{Short} & \multirow{2}*{Ant-v2} & 3.21 & 269.12 & 2661.56 & -1407.15 & 20.03 & \textbf{5359.85} \\
& & (1.97) & (224.66) & (1675.65) & (504.59) & (46.35) & \textbf{(129.28)} \\
& \multirow{2}*{Hopper-v2} & 286.53 & 3275.05 & 2508.76 & 287.38 & 474.21 & \textbf{3433.06} \\
& & (72.51) & (44.68) & (690.07) & (114.65) & (66.93) & \textbf{(96.89)} \\
& \multirow{2}*{HalfCheetah-v2} & 5.92 & 10480.33 & \textbf{10382.80} & 1225.35 & 3982.43 & \textbf{11073.88} \\
& & (23.06) & (202.27) & \textbf{(516.85)} & (162.88) & (433.75) & \textbf{(181.43)} \\
& \multirow{2}*{Walker2d-v2} & 222.21 & 3840.30 & 3999.14 & 328.97 & 348.36 & \textbf{5198.09} \\
& & (66.72) & (666.24) & (561.49) & (107.69) & (174.28) & \textbf{(225.50)} \\

\midrule

\multirow{8}*{Long} & \multirow{2}*{Ant-v2} & -115.10 & 215.38 & 2600.64 & -2552.99 & -0.53 & \textbf{4897.50} \\
& & (138.93) & (92.94) & (1229.27) & (419.44) & (5.07) & \textbf{(292.93)} \\
& \multirow{2}*{Hopper-v2} & 360.36 & 3015.96 & 3089.52 & 325.84 & 652.28 & \textbf{3447.64} \\
& & (118.06) & (408.08) & (433.23) & (50.25) & (92.41) & \textbf{(83.02)} \\
& \multirow{2}*{HalfCheetah-v2} & -115.56 & 5944.17 & 8591.38 & -571.35 & 4563.99 & \textbf{10880.76} \\
& & (35.01) & (3421.46) & (1048.08) & (374.64) & (568.19) & \textbf{(441.65)} \\
& \multirow{2}*{Walker2d-v2} & 251.11 & 3397.93 & 4221.37 & 284.27 & 497.27 & \textbf{4979.07} \\
& & (144.24) & (682.19) & (282.75) & (7.54) & (134.06) & \textbf{(166.95)} \\
\bottomrule
\end{tabular}}
\vskip -0.1in
\end{table*}

\begin{figure*}[t]
  \centering
  \includegraphics[width=0.98\textwidth]{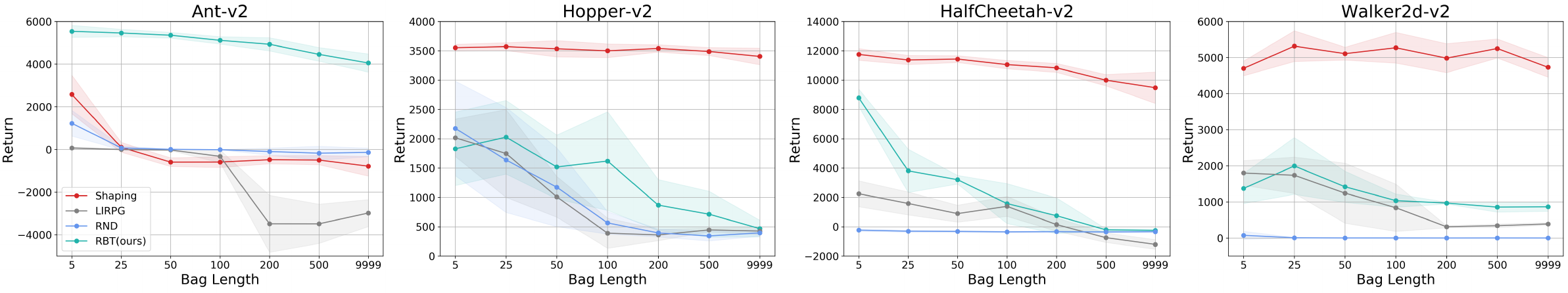}
  \caption{Comparison of proposed method and sparse reward baselines. The mean and standard deviation are displayed over 6 trials with different random seeds across a total of 1e6 time steps.}
  \label{app_fig:sparse_reward}
\end{figure*}

To ensure uniformity in the policy optimization process across all methodologies, each was subjected to 1,000,000 training iterations. For the proposed method, we initially collated a dataset comprising 10,000 time steps to pre-train the reward model. This model then underwent 100 pre-training iterations, a step deemed essential to adequately prepare the reward model before embarking on the principal policy learning phase. Following this initial warm-up period, the reward model was trained for 10 iterations after each new trajectory was incorporated. Moreover, to systematically gauge performance and progress, evaluations were carried out at intervals of every 5,000 time steps. The computational resources for these procedures were NVIDIA GeForce RTX 2080 Ti GPU clusters with 8GB of memory, dedicated to training and evaluating tasks.

\begin{figure*}[t]
  \centering
  \includegraphics[width=0.98\textwidth]{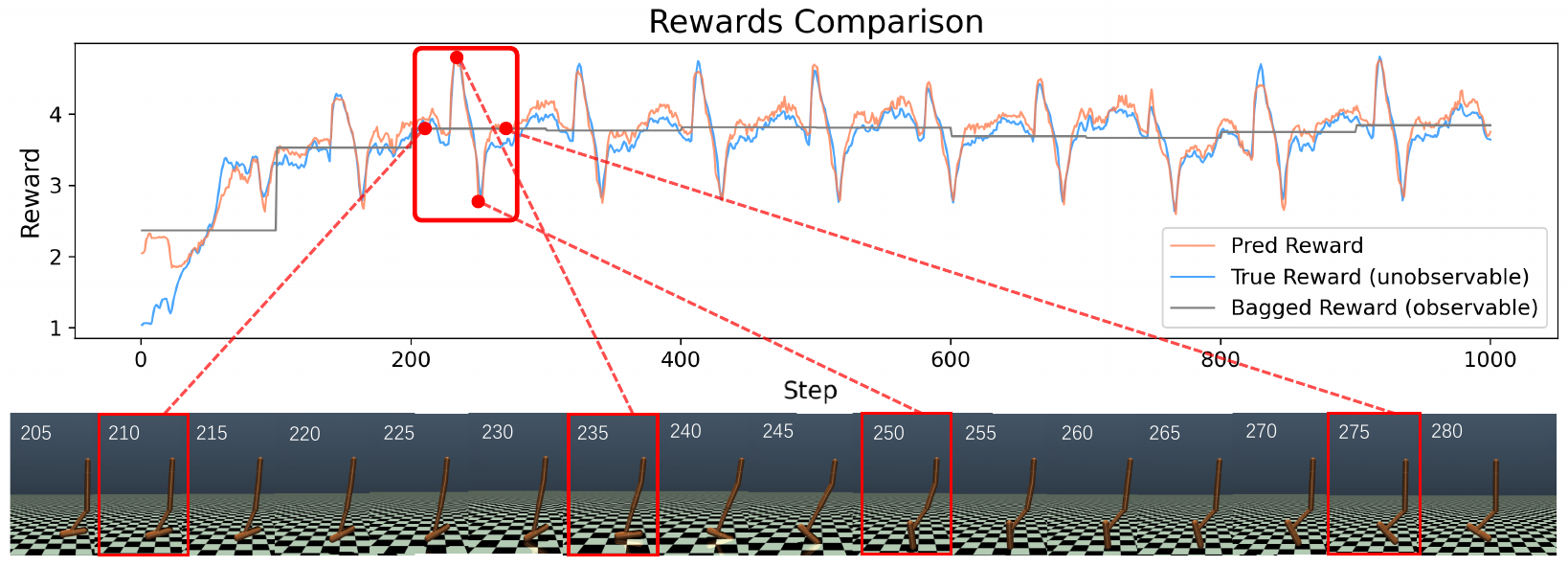}
  \caption{Rewards comparison and agent states in a trajectory with a bag length of 100 in the Hopper-v2 environment. The top graph compares predicted rewards against true rewards and aggregated bagged rewards.}
  \label{app_fig:reward_comparison}
\end{figure*}

\begin{figure*}[t]
  \centering
  \includegraphics[width=1\textwidth]{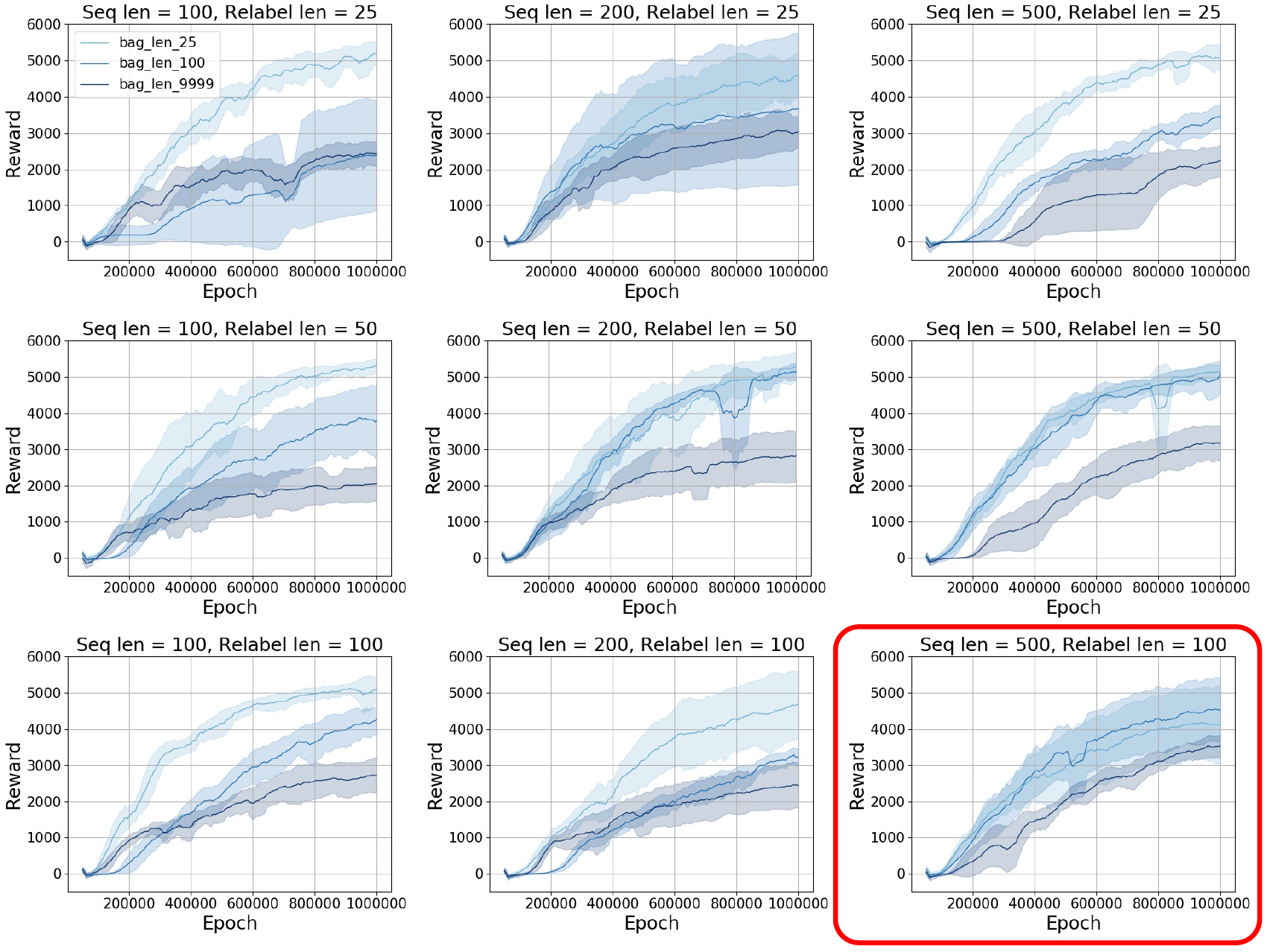}
  \caption{Learning curves on Ant-v2 with different length of input sequence in training (Seq len) and predict length during relabeling process (Relabel len), based on 3 independent runs with random initialization. Within each of the smaller graphs, the curves represent results from experiments with different bag lengths. Specifically, there are three bag lengths evaluated: 25, 100, and what is labeled as 9999, which we interpret as a proxy for trajectory feedback. The graph highlighted by the red box indicates our chosen parameter setting for the experiment, which is a input sequence length of 100 and a predict length of 500.}
  \label{fig:Architecture Sensitivity}
\end{figure*}

\section{Discussion on Sparse Reward}
The sparse reward setting presents significant challenges due to infrequent feedback, making it difficult for agents to effectively explore the environment and discover successful strategies. To address this challenge, various methods have been developed to enhance exploration. Reward shaping strategies \cite{ng1999policy, hu2020learning, tambwekar2019controllable} added rewards to actions in a way that guides the agent towards better policies without altering the original reward function. Curiosity-driven methods \cite{pathak2017curiosity, sekar2020planning} encouraged agents to explore the environment by visiting unseen states, potentially solving tasks with sparse rewards. Additionally, curriculum learning in RL \cite{florensa2018automatic, riedmiller2018learning} involved presenting an agent with a sequence of tasks with gradually increasing complexity, allowing the agent to eventually solve the initially given sparse reward task.

Due to the formal similarity between sparse rewards and trajectory feedback, both involving a single reward signal after a sequence, these methods can be directly applied to RLTF. However, the key difference lies in the nature of the rewards: sparse rewards typically focus on specific key points, making the reward Markovian, whereas RLTF evaluates the entire trajectory, indicating a Non-Markovian reward. Furthermore, sparse reward strategies address the exploration-exploitation trade-off inherent in sparse rewards, while RLTF focuses on distributing the evaluation of the entire trajectory across specific state-action pairs to guide learning. Thus, while these methods are applicable, they do not address the unique challenges posed by bagged rewards, as they fail to account for the reward structure within the bags. Therefore, they are not ideal solutions for RLTF~\cite{gangwani2020ircr, ren2021rrd}.

In Fig.~\ref{app_fig:sparse_reward}, we present additional results comparing our method with sparse reward baselines. Shaping~\cite{tesslerreward, hu2020learning} and LIRPG~\cite{zheng2018learning} are as described in the main paper. 
The Random Network Distillation (RND)~\cite{burda2019exploration} method introduces an exploration bonus for deep reinforcement learning by measuring the prediction error of a randomly initialized neural network, which is a widely recognized and utilized approach for addressing sparse reward challenges. We include it as a baseline in our comparison of sparse reward methods for a more comprehensive evaluation.

These results demonstrate that, despite the superficial similarity between sparse rewards and bagged rewards, the internal structure of the rewards differs, leading to suboptimal performance of sparse reward baselines in bagged reward environments. Conversely, as shown in the Section Compare with SOTA Methods of the main paper, our method performs well even in sparse reward environments, surpassing baselines specifically designed for sparse reward scenarios. This highlights the broad applicability and adaptability of our approach across different reward settings.

\section{Additional Experimental Results}
\label{app:Additional Experimental Results}
This section provides further analysis and insights through additional experiments to complement the main findings presented in our main paper.

\subsection{Experimental Result of Various-Length Reward Bags}
\label{sec_app_exp}
In Table~\ref{table:change_len}, we present the experiment results on various-length reward bags. The experiment depict in the table showcases the results of various methods applied across different environments with varying bag lengths of rewards, where bags are one next to another as in the definition of RLBR. This experiment reveals that longer bags tend to degrade the performance of most methods. However, our RBT method appears to be less sensitive to changes in bag length, maintaining robust performance even when the bag length is equal to the full trajectory. This result aligns with the result in the Section Compare with SOTA Methods in our main paper.

\subsection{Rewards Comparison in Hopper-v2 with Bagged Rewards}

Fig.~\ref{app_fig:reward_comparison} shows a comparison of predicted rewards, true rewards, and aggregated bagged rewards for a trajectory with a bag length of 100 in the Hopper-v2 environment. It shows how well the predicted rewards align with the true and bagged rewards over the course of the trajectory, highlighting the effectiveness of the reward model.

Beneath the figure, a series of images depicts a complete jump cycle by the agent, illustrating its motion sequence: mid-air, landing, jumping, and returning to mid-air. Red boxes highlight specific states that correspond to reward peaks and troughs, representing moments of maximum, minimum, and moderate rewards. 
In the Hopper-v2 environment, rewards consist of a constant ``healthy reward'' for operational integrity, a ``forward reward'' for progress in the positive x-direction, and a ``control cost'' for penalizing large actions. 
At peak reward instances, the agent is typically fully grounded in an optimal posture for forward leaping, which maximizes the ``forward reward'' through pronounced x-direction movement. Concurrently, it sustains the ``healthy reward'' and minimizes ``control cost'' through measured, efficient actions. This analysis underscores that the RBT can adeptly decode the environmental dynamics and the nuanced reward redistribution even under the setting of RLBR.

\subsection{Architecture Sensitivity}
\label{app:Architecture Sensitivity}

Fig.~\ref{fig:Architecture Sensitivity} illustrates the sensitivity of our architecture to different input sequence lengths during training (Seq len) and prediction lengths during reward relabeling (Relabel len) in the Ant-v2 environment. The learning curves, based on three independent runs with random initialization, show how varying these parameters affects the performance of the agent.

Although the configuration highlighted by the red box (Seq len of 500 and Relabel len of 100) demonstrates the best performance, the results also show that our proposed model is capable of learning effectively across various other configurations. This analysis underscores the importance of tuning input sequence length during training and prediction length during reward relabeling for optimal performance, while also demonstrating the ability of the RBT model to learn under different parameter settings, showcasing its flexibility and effectiveness in reinforcement learning tasks.

\end{document}